\documentclass[runningheads]{llncs}
\usepackage{etoolbox} 
\usepackage[frozencache=true,cachedir=minted-cache]{minted}
\makeatletter
\patchcmd{\ps@headings}{\rlap{\thepage}}{}{}{}
\patchcmd{\ps@headings}{\llap{\thepage}}{}{}{}
\makeatother
\usepackage{graphicx}
\usepackage{booktabs}
\usepackage{multirow}
\usepackage[table,xcdraw]{xcolor}
\usepackage{url}
\usepackage{dirtytalk}
\usepackage{float}

\begin{document}
\title{AI-UPV at IberLEF-2021 DETOXIS task: Toxicity Detection in Immigration-Related Web News Comments Using Transformers and Statistical Models}
\titlerunning{  }
%
\author{Angel Felipe Magnossão de Paula\orcidID{0000-0001-8575-5012} \and
Ipek Baris Schlicht\orcidID{0000-0002-5037-2203}
\institute{Universitat Politècnica de València, Spain \\ \{adepau, ibarsch\}@doctor.upv.es}
}
\authorrunning{  }
%
\maketitle              
%
{\let\thefootnote\relax\footnotetext{\textit{IberLEF 2021, September 2021, Málaga, Spain.}\\Copyright \textcopyright\ 2021 for this paper by its authors. Use permitted under Creative Commons License Attribution 4.0 International (CC BY 4.0).}} 

\begin{abstract}
This paper describes our participation in the DEtection of TOXicity in comments In Spanish (DETOXIS) shared task 2021 at the 3rd Workshop on Iberian Languages Evaluation Forum. The shared task is divided into two related classification tasks: (i) Task 1: toxicity detection and; (ii) Task 2: toxicity level detection. They focus on the xenophobic problem exacerbated by the spread of toxic comments posted in different online news articles related to immigration. One of the necessary efforts towards mitigating this problem is to detect toxicity in the comments. Our main objective was to implement an accurate model to detect xenophobia in comments about web news articles within the DETOXIS shared task 2021, based on the competition's official metrics: the F1-score for Task 1 and the Closeness Evaluation Metric (CEM) for Task 2. To solve the tasks, we worked with two types of machine learning models: (i) statistical models and (ii) Deep Bidirectional Transformers for Language Understanding (BERT) models. We obtained our best results in both tasks using BETO,  an BERT model trained on a big Spanish corpus. We obtained the 3rd place in Task 1 official ranking with the F1-score of 0.5996, and we achieved the 6th place in Task 2 official ranking with the CEM of 0.7142. Our results suggest: (i) BERT models obtain better results than statistical models for toxicity detection in text comments; (ii) Monolingual BERT models have an advantage over multilingual BERT models in toxicity detection in text comments in their pre-trained language.

\end{abstract}

\section{Introduction}

The increase in the number of news pages where the reader can openly discuss the articles has driven the dissemination of internet users' opinions through social media \cite{winter2015they,doi:10.1080/17512786.2020.1739550}. A survey carried out in the US by The Center for Media Engagement at the University of Texas at Austin states that most of the comments on news articles are posted by internet users who we call active-users or influencers \cite{stroud2016news}. They are highly active and generate huge amounts of data.

The imbalance in the amount of data generated by influencers and non-active users creates a distorted reality where influencers' opinions end up representing the opinion of all internet users to society \cite{10.1145/3366424.3383564}. This distorted reality can aggravate the existing social problems, as is the case with xenophobia, a heavy sense of aversion, or dread of people from other countries  \cite{xenophobia}.

In recent years, the problem with xenophobia has been exacerbated by the increase in the spread of toxic comments posted in different online news articles related to immigration \cite{blaya2019toward}. One of the first steps to mitigate the problem is to detect toxic comments regarding news articles \cite{davidson2017automated}. For this reason,  the Iberian Languages Evaluation Forum (IberLEF) proposed the DEtection of TOXicity in comments In Spanish (DETOXIS) shared task 2021 \cite{taule2021overview}. 

The DETOXIS shared task comprises Task 1 and Task 2, which are respectively toxicity detection and toxicity level detection. The two tasks are performed on comments posted in Spanish in response to different online news articles related to immigration. Task 1 is a binary classification problem where the objective is to classify a Spanish text comment as `toxic' or `not toxic'. Task 2 aims to classify the same comment but among four classes: `not toxic', `mildly toxic', `toxic', or `very toxic'. Table \ref{tab2:Comments_examples} displays examples of comments classified across all classes. The IberLEF's objective is to foster the research community towards innovative solutions for detecting xenophobia on social media platforms \cite{taule2021overview}.

\begin{table}[]
\centering
\caption{Comments examples}
\label{tab2:Comments_examples}
\begin{tabular}{@{}|c|c|l|@{}}
\toprule
Toxicity & Toxicity\_level & \multicolumn{1}{c|}{New’s comment} \\ \midrule
not toxic & not toxic & Proximamente en su barrio \\ \midrule
\multirow{3}{*}{toxic} & mildly toxic & Vienen a pagarnos las pensiones \\ \cmidrule(l){2-3} 
 & toxic & \begin{tabular}[c]{@{}l@{}}asi me gusta, que se maten entre ellos y en alta mar. Mas\\ inmigrantes asi porfavor\end{tabular} \\ \cmidrule(l){2-3} 
 & very toxic & \begin{tabular}[c]{@{}l@{}}A esosmoros hay que echarlos pero ya.O los politicos \\ hacen algo o la gente tendra que "actuar"\end{tabular} \\ \bottomrule
\end{tabular}
\end{table}

Nowadays, the detection of toxicity in comments is done with Machine Learning (ML)  models, especially deep learning models, which require large amounts of annotated datasets for robust predictions \cite{8005992}. However, labeling toxicity is a challenging and time-consuming task that requires many annotators to avoid bias, and the annotators should be aware of social and cultural contexts \cite{risch2018delete,korencic-etal-2021-block}.

Our main goal was to implement an accurate model to detect xenophobic comments on web news articles within the DETOXIS shared task 2021, using the competition's official metrics. We decided to solve the problem by applying models that can learn using only a small amount of data, which can be done with statistical models and most advanced pre-trained deep learning models. Roughly speaking, there are two types of statistical models: Generative and Discriminative \cite{jebara2012machine}. We chose to use one of each type. Thus, we tried a Naive Bayes (Generative) and a Maximum Entropy (Discriminative) model. Among the most advanced and highly effective deep learning models is Deep Bidirectional Transformers for Language Understanding (BERT), which comes with its parameters pre-trained in an unsupervised manner in a large corpus \cite{devlin2018BERT}. Therefore, it only needs a tuned train that can be run on a small set of data, which suits our problem. Our source code is publicly available\footnote{\url{https://github.com/AngelFelipeMP/Machine-Learning-Tweets-Classification}}

The work's main contribution is to help in the effort to improve the results in the identification of toxic comments in news articles related to immigration. Unlike the vast majority of works \cite{10.1145/3369869}, we use ML models that can tackle the xenophobia detection problem having only little data available. The second contribution is to build ML models and find their best configuration to deal not only with the classification of news articles as `toxic' and `not toxic', but also to infer the toxicity level of the comments into `not toxic', `mildly toxic', `toxic', or `very toxic'. As far as we know, there are few works in the literature in which the solution model tries to infer the toxicity level of the comments posted in the news related to immigration. On the DETOXIS official ranking, we obtained the 3rd place in Task 1 with the F1-score of 0.5996, and we achieved the 6th place in Task 2 with the CEM of 0.7142.

The article is organized as follows: Section 2 contains the methodology with fundamental concepts; Section 3 describes the experiments; Section 4 contains the results and discussions, and and Section 5 draws some conclusion and future work.

\section{Methodology}

This section explains the data structure, the evaluation metrics, and the ML models applied to solve our classification problems. In addition, the text representation used to encode the text comments.

\subsection{Dataset}

The DETOXIS shared task organization granted its participants the NewsCom-TOX dataset \cite{taule2021overview} divided into train set and test set where text data are in Spanish. The train set consists of 3463 instances, and the test set consists of 891 instances. Both sets have as main labels: (i) `Comment\_id' and (ii) `Comment'; but only the train set has the labels: (iii) `Toxicity' and (iv) `Toxicity\_level', respectively for Task 1 and Task 2. The `Comment\_id' is a unique reference number assigned to each instance within the NewsCom-TOX dataset. The `Comment' label is a text message posted in response to a Spanish online news article from different sources such as El Mundo, NIUS, ABC, etc., or discussion forums like Menéame. Moreover, `Toxicity' labels the comment for a particular instance between `toxic' or `not toxic' and the `Toxicity\_level' label classifies the same comment as `not toxic', `mildly toxic', `toxic', or `very toxic'. Table \ref{tab1:Data_distribution} shows the label's distribution for `Toxicity' and `Toxicity\_level'. We can see that the labels are unbalanced in both cases.

\begin{table}[]
\centering
\caption{Data distribution}
\label{tab1:Data_distribution}
\begin{tabular}{@{}|c|c|c|c|@{}}
\toprule
\multicolumn{2}{|c|}{Toxicity (Task 1)} & \multicolumn{2}{c|}{Toxicity\_level (Task 2)} \\ \midrule
Label & Number of instances & Label & Number of instances \\ \midrule
not toxic & 2316 & not toxic & 2317 \\ \midrule
\multirow{3}{*}{toxic} & \multirow{3}{*}{1147} & mildly toxic & 808 \\ \cmidrule(l){3-4} 
 &  & toxic & 269 \\ \cmidrule(l){3-4} 
 &  & very toxic & 69 \\ \bottomrule
\end{tabular}
\end{table}

The data annotation process was carried out by four annotators where two were linguists experts, and two were trained linguistic students. Three of them labeled all news article comments in parallel. Once they finished, an inter-annotator agreement test was executed. When a disagreement happens, the three annotators plus the senior annotator reviewed it in order to achieve accordance with the final label \cite{taule2021overview}. In Table \ref{tab2:Comments_examples}, we can see examples from the DETOXIS train set of comments and its labels attributed by the annotators for `Toxicity' and `Toxicity\_level'.

\begin{figure}
     \centering
    \includegraphics[scale=0.05]{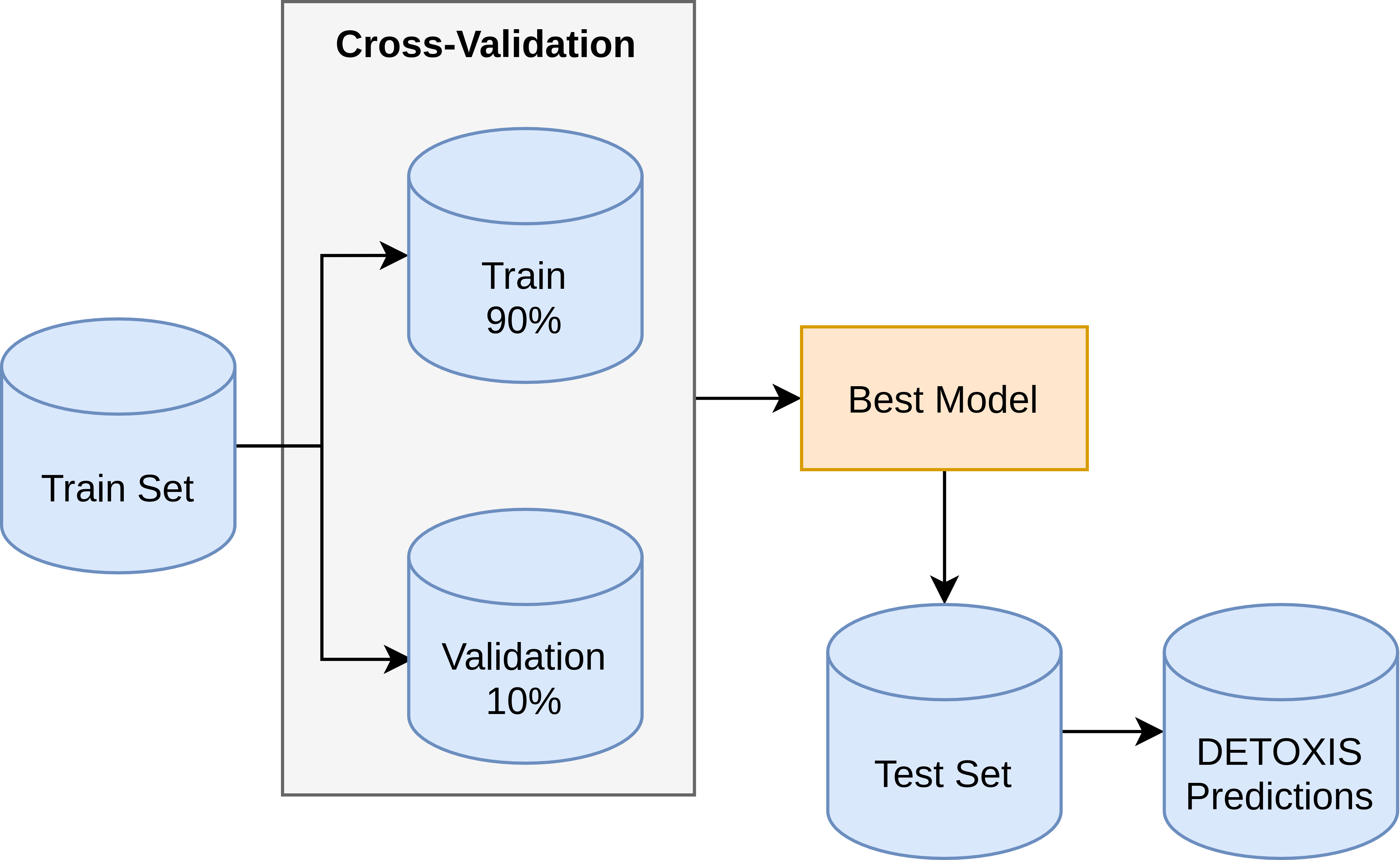}
    \caption{Workflow.} 
    \label{fig1:Data_pipeline}
\end{figure}

Next we explain how we used the data during the project development in both tasks. First, we applied 10-fold cross-validation in the train set to find the best ML model. After that, we trained the selected model in the whole train set. Subsequently, we applied the selected model to make predictions on the official test set, as shown in Figure \ref{fig1:Data_pipeline}. These predictions were submitted to the DETOXIS shared task 2021.

\subsection{Evaluation metrics}

Because the train set is unbalanced, as we can see in Table \ref{tab1:Data_distribution}, we selected evaluation metrics that are able to fairly evaluate ML models in this circumstance.  For Task 1, we adopted Accuracy, Recall, Precision, and F1-score, which was the DETOXIS official evaluation metric for Task 1. For Task 2, we adopted Accuracy, F1-macro, F1-weighted, Recall, Precision, and CEM \cite{amigo2020effectiveness}, the DETOXIS official evaluation metric for Task 2. We used the DETOXIS official metrics as performance measures to rank and select the best ML models during the cross-validation process for Task 1 and Task 2.

\subsection{Models}

There are two types of statistical models: the Generative models and the Discriminative models \cite{jebara2012machine}. We used one model from each kind. We adopted the Naive Bayes (Generative) model and the Maximum Entropy (Discriminative) model. Among the Transformers models, we decided to use the BERT models, one of the most advanced and highly effective Transformer models. They come with their parameters pre-trained in an unsupervised manner in a large corpus \cite{devlin2018BERT}. Therefore, they only need a supervised fine-tune train in the downstream task that can be run on a small set of data.  We adopted: (i) the BETO model, a BERT model trained on a big unannotated Spanish corpus composed of three billion tokens \cite{canete2020spanish}; and (ii) the mBERT, a BERT model pre-trained on the top 102 languages with the most extensive Wikipedia corpus. However, the balance among the language in the corpus was not perfect. For example, the English partition of the corpus was 1000 bigger than the Icelandic partition \cite{Charles2013}.

\subsection{Text representation}

To represent our text data in a way that the statistical models could handle it, we used two encode methods: (i) Bag of Words (BOW) \cite{zhang2010understanding}; and (ii) Term Frequency - Inverse Document Frequency (TF-IDF) \cite{pimpalkar2020influence}. The BOW represents a text comment by a unidimensional vector whose length is the size of the training vocabulary. In this case, each column of this vector contains the number of times a particular word from the vocabulary appears in the specific comment. The TF-IDF representation for each text comment is also a flat vector with the size of the training vocabulary. However, the value for each word on the vector follows the well-known TF-IDF calculation \cite{pimpalkar2020influence}.

\section{Experiments}

This section explains the environment setup, the data preprocessing, and statistical models' feature extraction.  Furthermore, the section also contains explanations for the 10-fold cross-validation process and how we selected the model to make predictions on the DETOXIS test set, which we submitted as our final results to the competition.

\subsection{Environment setup}

For code purposes, we used python 3.7.10. As a code editor/machine, we used Google collaborator, also known as google colab. We made this choice because google provides free TPU's, which we used to train our BERT models. The main python libraries that we used were: (i) NumPy 1.19.5 to work with matrix, (ii) Pandas 1.1.5 to handle and visualize data, (iii) Spacy 2.2.4 and (iv) the Natural Language Toolkit (NLTK) 3.2.5 for natural language transformations, (v) Pytorch 1.6.0, and (vi) Transformers 3.0.0 to actually implement the BERT models. In addition, we used (vii) Sklearn 0.22.2 to implement the statistical models.

\subsection{Preprocessing}

For both tasks, we only preprocess the data for the statistical models. The preprocessing step was carried out on the text data from the train and test sets. We used the built-in python model for Regular Expression (ReEx) and the NLTK python library.  Applying ReEx, we removed stock market tickers, old-style retweet text, hashtags, hyperlinks and changed the numbers to the tag \say{\textless number\textgreater}.  We employed the NLTK on the text comments to remove stop-words, stem and tokenize the words.

\subsection{Feature extraction}

The feature extraction process was executed to focus on achieving good results with the statistical models. These models' performance is susceptible to their input features \cite{nelson2009accelerated}. Hence, after preprocessing the datasets for the statistical models, we executed the feature extraction process to create good input features. We encode the text comments in two different manners: (i) BOW \cite{zhang2010understanding}; and (ii) TF-IDF \cite{pimpalkar2020influence}.

The two proposed encode methods are based on word occurrences, and unfortunately, they completely ignore the relative position information of the words in the comments. Therefore, we lose the information about the local ordering of the words. In order to mitigate this problem and preserve some of the local word ordering information, we increase the vocabulary by extracting 2-grams and 3-grams from the text comments despite dimensional increasing.

\subsection{Cross-validation}

\begin{figure}
     \centering
    \includegraphics[scale=0.05]{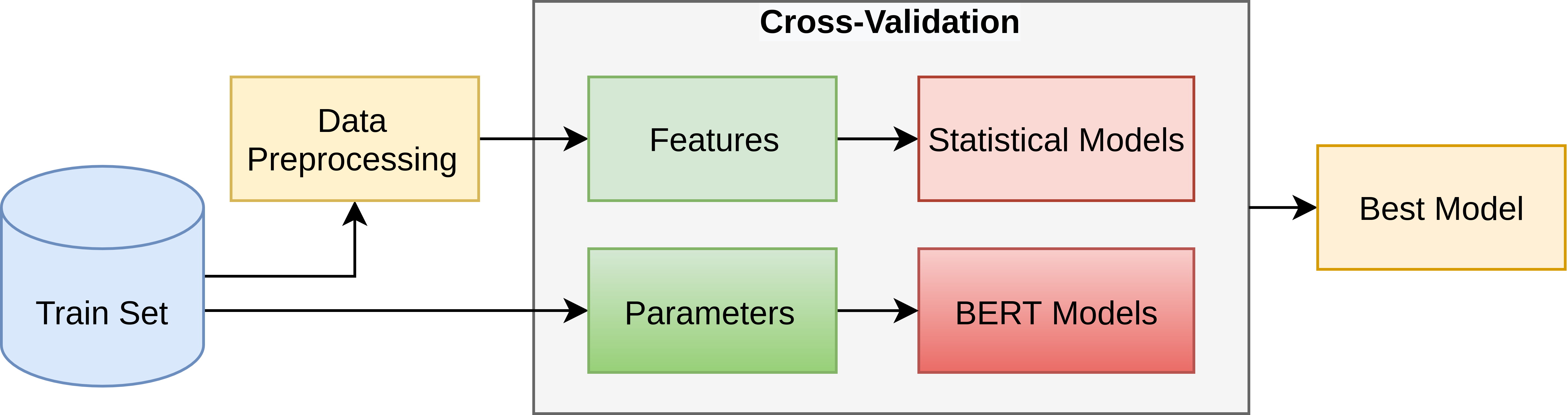}
    \caption{General diagram ML models cross-validation.} 
    \label{fig2:General_diagram_ML_models_cross-validation}
\end{figure}

The cross-validation process was performed on the train set aiming to find the best ML model to make the prediction on the DETOXIS test set. We can see the summary of our cross-validation process in Figure \ref{fig2:General_diagram_ML_models_cross-validation}. During the cross-validation, each statistical model received different input features, and the BERT models tried different hyper-parameters.

\begin{figure}
     \centering
    \includegraphics[scale=0.05]{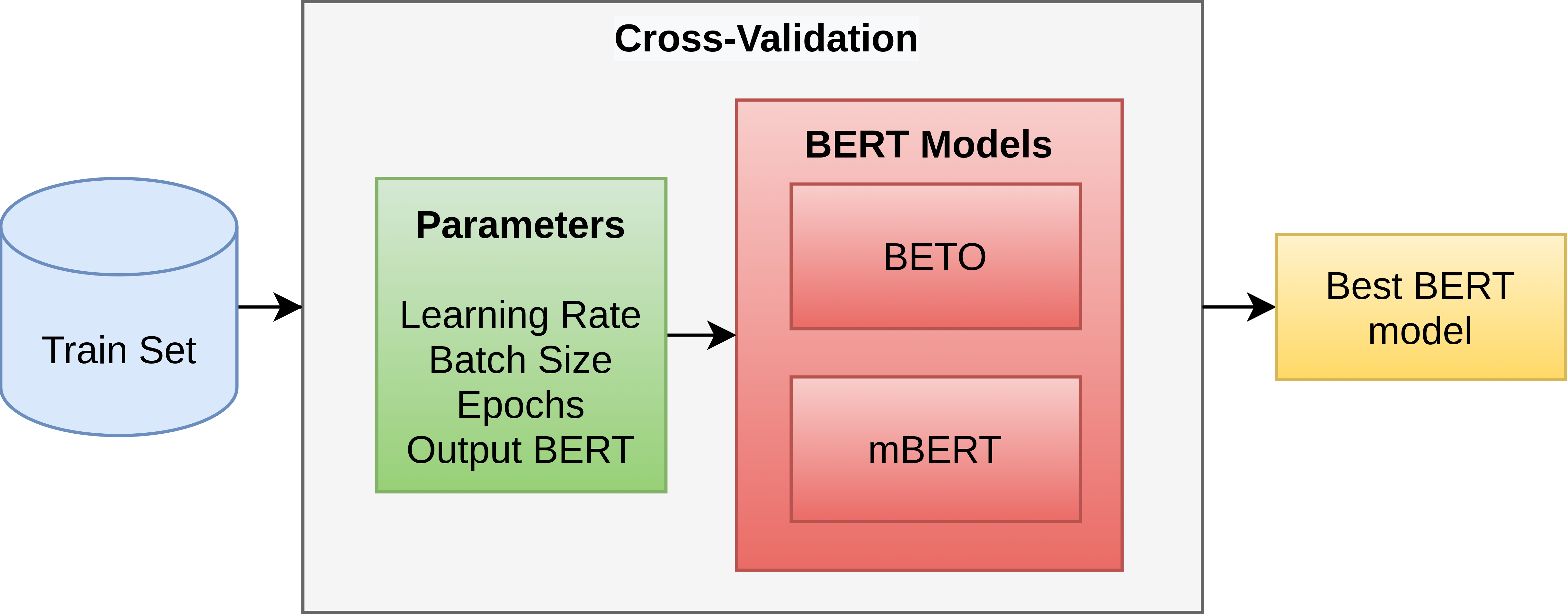}
    \caption{Cross-validation BERT models.} 
    \label{fig3:Cross-validation_BERT_models}
\end{figure}

In Figure \ref{fig3:Cross-validation_BERT_models},  we can see the cross-validation process that focuses on the BERT models. We tried different combinations for the Output BERT, Learning Rate, Batch Size, and Epochs. The BERT models are composed of a pre-trained model plus a linear layer at the top which receives the output of the pre-trained BERT model. We have two different options for the Output BERT: (i) the sequence of hidden states at the output of the last layer which we performed a mean pooling and max pooling operation and concatenated them into a unified unidimensional vector that we called `hidden'; (ii) the pooler of the last layer’s hidden state of the first token of the sequence further processed by a linear layer and a tanh activation function that we called `pooler'. For the Learning Rate, we tried 1E-5, 3E-5, and  5E-5. For the Batch Size, we tried 8, 16, 32, and 64. The number of Epochs was from 1 to 20.

\begin{figure}[]
     \centering
    \includegraphics[width=\textwidth]{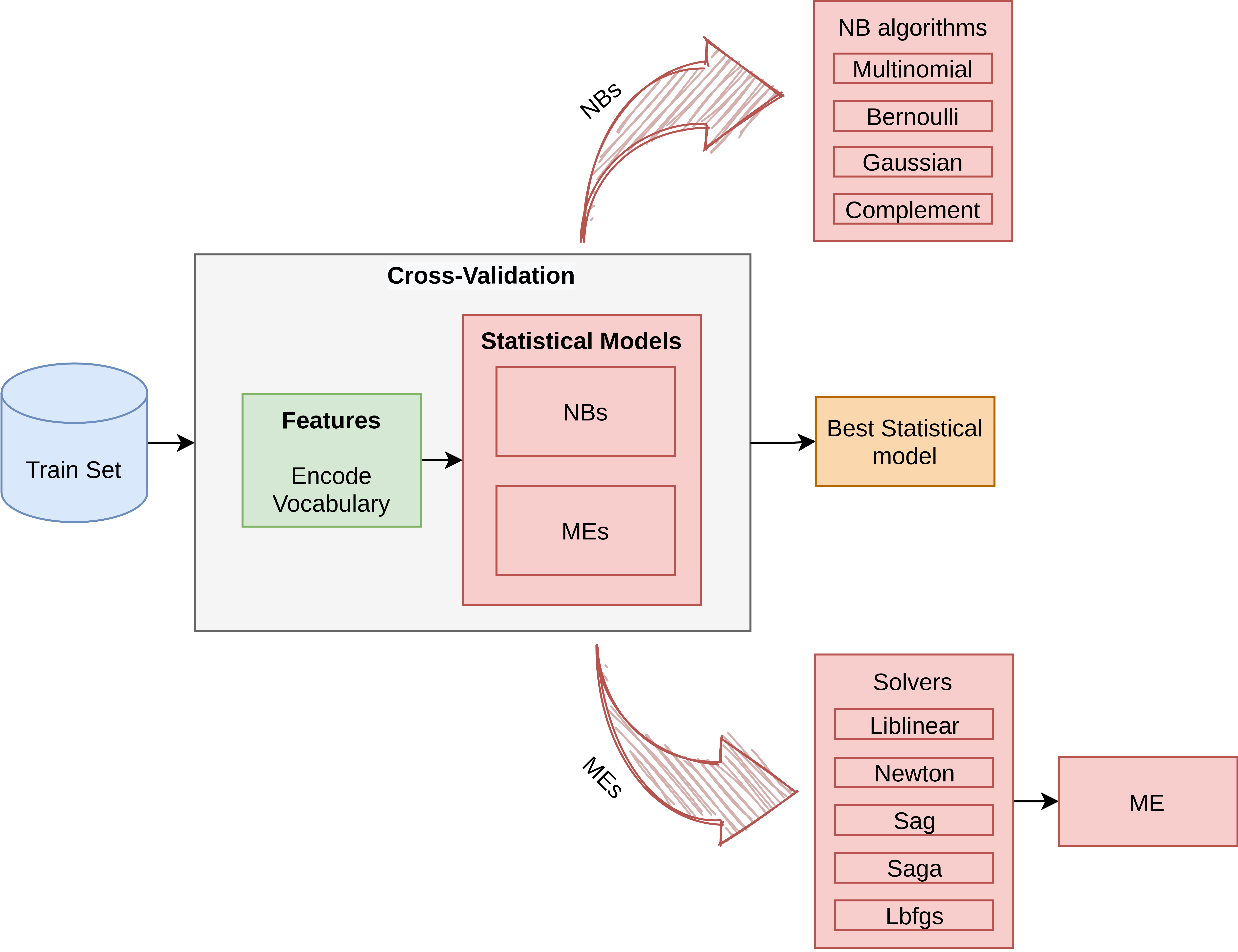}
    \caption{Cross-validation statistical models.} 
    \label{fig4:Cross-validation_statistical_models}
\end{figure}

Figure \ref{fig4:Cross-validation_statistical_models} illustrates the cross-validation process for the statistical models. We can see that we tried four algorithm versions of the Naive Bayes (NB) model: the Multinomial, the Bernoulli, the Gaussian, and the Complement ones. On the other hand, we tried only the original version of the Maximum Entropy (ME) model but with different solvers: the liblinear, newton, sag, saga, and lbfgs. We call solvers the algorithms used in the optimization problem. We tried different vocabulary sizes for all statistical models using different n-grams combinations and the two encode methods: BOW and TF-IDF.

Tables \ref{tab3:Cross-validation_ME_models_Task_1}, \ref{tab4:Cross-validation_NB_models_Task_1}, \ref{tab5:Cross-validation_ME_models_Task_2}, and \ref{tab6:Cross-validation_NB_ models_Task_2} show the results of the 10-fold cross-validation process for the statistical models. Tables \ref{tab3:Cross-validation_ME_models_Task_1} and \ref{tab4:Cross-validation_NB_models_Task_1}, in this sequence, show the results of the ME model and the NB model for Task 1. Tables \ref{tab5:Cross-validation_ME_models_Task_2} and \ref{tab6:Cross-validation_NB_ models_Task_2} respectively show the results of the ME model and the NB model for Task 2. The first column in Tables \ref{tab3:Cross-validation_ME_models_Task_1} and \ref{tab4:Cross-validation_NB_models_Task_1} shows the solver, and the first column in Tables \ref{tab5:Cross-validation_ME_models_Task_2} and \ref{tab6:Cross-validation_NB_ models_Task_2} shows the NB algorithm. All the other columns have the same meaning in the four tables.

Thus, the tables' third column shows the n-grams used as a vocabulary where the numbers within the parentheses are the lower and upper limits of the n-gram word range used. The n-gram range of (1, 1) means only 1-grams, (1, 2) means 1-grams and 2-grams, and (1, 3) means 1-grams, 2-grams, and 3-grams. The rest of the columns have the evaluation metrics for each group of the selected parameters. For Task 1, the evaluation metrics are Accuracy, F1-score, Recall, and Precision, and for Task 2, the evaluation metrics are Accuracy, F1-macro, F1-weighted, Recall, Precision, and CEM.

\begin{table}[H]
\centering
\caption{Cross-validation ME models for Task 1}
\label{tab3:Cross-validation_ME_models_Task_1}
\scalebox{0.95}{
\begin{tabular}{@{}ccccccc@{}}
\toprule
Solver & Encoder & Vocabulary & Accuracy & F1-score & Recall & Precision \\ \midrule
\multirow{6}{*}{Liblinear} & \multirow{3}{*}{TF-IDF} & (1,1)-grams & 0.7112 & 0.3031 & 0.2021 & 0.6882 \\ \cmidrule(l){3-7} 
 &  & (1,2)-grams & 0.6976 & 0.1546 & 0.0949 & 0.8786 \\ \cmidrule(l){3-7} 
 &  & (1,3)-grams & 0.6893 & 0.1036 & 0.0635 & 0.6500 \\ \cmidrule(l){2-7} 
 & \multirow{3}{*}{BOW} & (1,1)-grams & 0.6994 & 0.4652 & 0.3993 & 0.5649 \\ \cmidrule(l){3-7} 
 &  & (1,2)-grams & 0.7118 & 0.4353 & 0.3434 & 0.6060 \\ \cmidrule(l){3-7} 
 &  & (1,3)-grams & \textbf{0.7126} & 0.4101 & 0.3128 & 0.6188 \\ \midrule
\multirow{6}{*}{Newton} & \multirow{3}{*}{TF-IDF} & (1,1)-grams & 0.7115 & 0.3043 & 0.2030 & 0.6890 \\ \cmidrule(l){3-7} 
 &  & (1,2)-grams & 0.6979 & 0.1547 & 0.0949 & \textbf{0.8928} \\ \cmidrule(l){3-7} 
 &  & (1,3)-grams & 0.6893 & 0.1036 & 0.0635 & 0.6500 \\ \cmidrule(l){2-7} 
 & \multirow{3}{*}{BOW} & (1,1)-grams & 0.6991 & 0.4645 & 0.3984 & 0.5647 \\ \cmidrule(l){3-7} 
 &  & (1,2)-grams & 0.7106 & 0.4331 & 0.3416 & 0.6030 \\ \cmidrule(l){3-7} 
 &  & (1,3)-grams & 0.7132 & 0.4106 & 0.3128 & 0.6212 \\ \midrule
\multirow{6}{*}{Sag} & \multirow{3}{*}{TF-IDF} & (1,1)-grams & 0.7115 & 0.3043 & 0.2030 & 0.6890 \\ \cmidrule(l){3-7} 
 &  & (1,2)-grams & 0.6979 & 0.1547 & 0.0949 & \textbf{0.8928} \\ \cmidrule(l){3-7} 
 &  & (1,3)-grams & 0.6893 & 0.1036 & 0.0635 & 0.6500 \\ \cmidrule(l){2-7} 
 & \multirow{3}{*}{BOW} & (1,1)-grams & 0.7002 & \textbf{0.4679} & \textbf{0.4019} & 0.5670 \\ \cmidrule(l){3-7} 
 &  & (1,2)-grams & 0.7141 & 0.4427 & 0.3512 & 0.6110 \\ \cmidrule(l){3-7} 
 &  & (1,3)-grams & 0.7155 & 0.4168 & 0.3189 & 0.6248 \\ \midrule
\multirow{6}{*}{Saga} & \multirow{3}{*}{TF-IDF} & (1,1)-grams & 0.7112 & 0.3031 & 0.2021 & 0.6882 \\ \cmidrule(l){3-7} 
 &  & (1,2)-grams & 0.6979 & 0.1562 & 0.0957 & 0.8800 \\ \cmidrule(l){3-7} 
 &  & (1,3)-grams & 0.6893 & 0.1036 & 0.0635 & 0.6500 \\ \cmidrule(l){2-7} 
 & \multirow{3}{*}{BOW} & (1,1)-grams & 0.6985 & 0.4652 & 0.4002 & 0.5629 \\ \cmidrule(l){3-7} 
 &  & (1,2)-grams & 0.7135 & 0.4432 & 0.3521 & 0.6102 \\ \cmidrule(l){3-7} 
 &  & (1,3)-grams & 0.7181 & 0.4246 & 0.3242 & 0.6337 \\ \midrule
\multirow{6}{*}{Lbfgs} & \multirow{3}{*}{TF-IDF} & (1,1)-grams & 0.7115 & 0.3043 & 0.2030 & 0.6890 \\ \cmidrule(l){3-7} 
 &  & (1,2)-grams & 0.6979 & 0.1547 & 0.0949 & \textbf{0.8928} \\ \cmidrule(l){3-7} 
 &  & (1,3)-grams & 0.6893 & 0.1036 & 0.0635 & 0.6500 \\ \cmidrule(l){2-7} 
 & \multirow{3}{*}{BOW} & (1,1)-grams & 0.6991 & 0.4645 & 0.3984 & 0.5647 \\ \cmidrule(l){3-7} 
 &  & (1,2)-grams & 0.7106 & 0.4331 & 0.3416 & 0.6030 \\ \cmidrule(l){3-7} 
 &  & (1,3)-grams & 0.7132 & 0.4106 & 0.3128 & 0.6212 \\ \bottomrule
\end{tabular}
}
\end{table}

\begin{table}[]
\centering
\caption{Cross-validation NB models for Task 1}
\label{tab4:Cross-validation_NB_models_Task_1}
\scalebox{0.95}{
\begin{tabular}{@{}ccccccc@{}}
\toprule
NB Algorithm & Encoder & Vocabulary & Accuracy & F1-score & Recall & Precision \\ \midrule
\multirow{6}{*}{Multinomial} & \multirow{3}{*}{TF-IDF} & (1,1)-grams & \textbf{0.6933} & 0.1703 & 0.1062 & \textbf{0.7282} \\ \cmidrule(l){3-7} 
 &  & (1,2)-grams & 0.6878 & 0.0995 & 0.0609 & 0.7167 \\ \cmidrule(l){3-7} 
 &  & (1,3)-grams & 0.6843 & 0.0805 & 0.0487 & 0.5500 \\ \cmidrule(l){2-7} 
 & \multirow{3}{*}{BOW} & (1,1)-grams & 0.6685 & 0.4868 & 0.4821 & 0.4960 \\ \cmidrule(l){3-7} 
 &  & (1,2)-grams & 0.6480 & 0.5232 & 0.5868 & 0.4736 \\ \cmidrule(l){3-7} 
 &  & (1,3)-grams & 0.5795 & \textbf{0.5355} & 0.7289 & 0.4238 \\ \midrule
\multirow{6}{*}{Bernoulli} & \multirow{3}{*}{TF-IDF} & (1,1)-grams & 0.6674 & 0.3344 & 0.2574 & 0.4979 \\ \cmidrule(l){3-7} 
 &  & (1,2)-grams & 0.6524 & 0.1910 & 0.1274 & 0.4297 \\ \cmidrule(l){3-7} 
 &  & (1,3)-grams & 0.6529 & 0.1765 & 0.1160 & 0.4168 \\ \cmidrule(l){2-7} 
 & \multirow{3}{*}{BOW} & (1,1)-grams & 0.6674 & 0.3344 & 0.2574 & 0.4979 \\ \cmidrule(l){3-7} 
 &  & (1,2)-grams & 0.6524 & 0.1910 & 0.1274 & 0.4297 \\ \cmidrule(l){3-7} 
 &  & (1,3)-grams & 0.6529 & 0.1765 & 0.1160 & 0.4168 \\ \midrule
\multirow{6}{*}{Gaussian} & \multirow{3}{*}{TF-IDF} & (1,1)-grams & 0.4730 & 0.4192 & 0.5831 & 0.3294 \\ \cmidrule(l){3-7} 
 &  & (1,2)-grams & 0.5287 & 0.3918 & 0.4733 & 0.3386 \\ \cmidrule(l){3-7} 
 &  & (1,3)-grams & 0.5307 & 0.3961 & 0.4794 & 0.3418 \\ \cmidrule(l){2-7} 
 & \multirow{3}{*}{BOW} & (1,1)-grams & 0.4675 & 0.4282 & 0.6102 & 0.3317 \\ \cmidrule(l){3-7} 
 &  & (1,2)-grams & 0.5223 & 0.4068 & 0.5090 & 0.3425 \\ \cmidrule(l){3-7} 
 &  & (1,3)-grams & 0.5249 & 0.4084 & 0.5099 & 0.3442 \\ \midrule
\multirow{6}{*}{Complement} & \multirow{3}{*}{TF-IDF} & (1,1)-grams & 0.6604 & 0.4083 & 0.3800 & 0.4633 \\ \cmidrule(l){3-7} 
 &  & (1,2)-grams & 0.6785 & 0.3255 & 0.2648 & 0.4845 \\ \cmidrule(l){3-7} 
 &  & (1,3)-grams & 0.6835 & 0.3378 & 0.2727 & 0.5071 \\ \cmidrule(l){2-7} 
 & \multirow{3}{*}{BOW} & (1,1)-grams & 0.6234 & 0.5165 & 0.6156 & 0.4472 \\ \cmidrule(l){3-7} 
 &  & (1,2)-grams & 0.5928 & 0.5216 & 0.6749 & 0.4263 \\ \cmidrule(l){3-7} 
 &  & (1,3)-grams & 0.5215 & 0.5256 & \textbf{0.8004} & 0.3915 \\ \bottomrule
\end{tabular}
}
\end{table}

\begin{table}[H]
\centering
\caption{Cross-validation ME models for Task 2}
\label{tab5:Cross-validation_ME_models_Task_2}
\begin{tabular}{@{}ccccccccc@{}}
\toprule
Solver & Encoder & Vocabulary & Accuracy & F1-macro & F1-weighted & Recall & Precision & CEM \\ \midrule
\multirow{6}{*}{Liblinear} & \multirow{3}{*}{TF-IDF} & (1,1)-grams & 0.6826 & 0.2363 & 0.5753 & 0.2691 & 0.2798 & 0.7070 \\ \cmidrule(l){3-9} 
 &  & (1,2)-grams & 0.6809 & 0.2233 & 0.5610 & 0.2629 & 0.2693 & 0.6972 \\ \cmidrule(l){3-9} 
 &  & (1,3)-grams & 0.6797 & 0.2214 & 0.5585 & 0.2616 & 0.2649 & 0.6923 \\ \cmidrule(l){2-9} 
 & \multirow{3}{*}{BOW} & (1,1)-grams & 0.6526 & 0.3236 & 0.6034 & 0.3129 & 0.4473 & 0.6831 \\ \cmidrule(l){3-9} 
 &  & (1,2)-grams & 0.6722 & 0.3070 & 0.6038 & 0.3059 & 0.4539 & 0.7018 \\ \cmidrule(l){3-9} 
 &  & (1,3)-grams & 0.6780 & 0.2944 & 0.5998 & 0.2995 & 0.4418 & \textbf{0.7080} \\ \midrule
\multirow{6}{*}{Newton} & \multirow{3}{*}{TF-IDF} & (1,1)-grams & 0.6826 & 0.2509 & 0.5870 & 0.2767 & 0.3199 & 0.7067 \\ \cmidrule(l){3-9} 
 &  & (1,2)-grams & \textbf{0.6846} & 0.2302 & 0.5691 & 0.2675 & 0.2934 & 0.7041 \\ \cmidrule(l){3-9} 
 &  & (1,3)-grams & 0.6829 & 0.2267 & 0.5643 & 0.2652 & 0.2649 & 0.6977 \\ \cmidrule(l){2-9} 
 & \multirow{3}{*}{BOW} & (1,1)-grams & 0.6465 & \textbf{0.3587} & \textbf{0.6125} & \textbf{0.3367} & \textbf{0.4942} & 0.6827 \\ \cmidrule(l){3-9} 
 &  & (1,2)-grams & 0.6682 & 0.3176 & 0.6079 & 0.3117 & 0.4504 & 0.6984 \\ \cmidrule(l){3-9} 
 &  & (1,3)-grams & 0.6740 & 0.2997 & 0.6028 & 0.3019 & 0.4303 & 0.7022 \\ \midrule
\multirow{6}{*}{Sag} & \multirow{3}{*}{TF-IDF} & (1,1)-grams & 0.6826 & 0.2509 & 0.5870 & 0.2767 & 0.3199 & 0.7067 \\ \cmidrule(l){3-9} 
 &  & (1,2)-grams & \textbf{0.6846} & 0.2302 & 0.5691 & 0.2675 & 0.2934 & 0.7041 \\ \cmidrule(l){3-9} 
 &  & (1,3)-grams & 0.6829 & 0.2267 & 0.5643 & 0.2652 & 0.2649 & 0.6977 \\ \cmidrule(l){2-9} 
 & \multirow{3}{*}{BOW} & (1,1)-grams & 0.6460 & 0.3393 & 0.6107 & 0.3251 & 0.4386 & 0.6824 \\ \cmidrule(l){3-9} 
 &  & (1,2)-grams & 0.6690 & 0.3101 & 0.6077 & 0.3073 & 0.4306 & 0.6997 \\ \cmidrule(l){3-9} 
 &  & (1,3)-grams & 0.6742 & 0.2919 & 0.6021 & 0.2977 & 0.3976 & 0.7039 \\ \midrule
\multirow{6}{*}{Saga} & \multirow{3}{*}{TF-IDF} & (1,1)-grams & 0.6826 & 0.2509 & 0.5870 & 0.2767 & 0.3199 & 0.7067 \\ \cmidrule(l){3-9} 
 &  & (1,2)-grams & \textbf{0.6846} & 0.2302 & 0.5691 & 0.2675 & 0.2934 & 0.7041 \\ \cmidrule(l){3-9} 
 &  & (1,3)-grams & 0.6829 & 0.2267 & 0.5643 & 0.2652 & 0.2649 & 0.6977 \\ \cmidrule(l){2-9} 
 & \multirow{3}{*}{BOW} & (1,1)-grams & 0.6459 & 0.3380 & 0.6102 & 0.3241 & 0.4387 & 0.6825 \\ \cmidrule(l){3-9} 
 &  & (1,2)-grams & 0.6699 & 0.3145 & 0.6077 & 0.3101 & 0.4526 & 0.7015 \\ \cmidrule(l){3-9} 
 &  & (1,3)-grams & 0.6763 & 0.2902 & 0.6027 & 0.2974 & 0.4028 & 0.7053 \\ \midrule
\multirow{6}{*}{Lbfgs} & \multirow{3}{*}{TF-IDF} & (1,1)-grams & 0.6826 & 0.2509 & 0.5870 & 0.2767 & 0.3199 & 0.7067 \\ \cmidrule(l){3-9} 
 &  & (1,2)-grams & \textbf{0.6846} & 0.2302 & 0.5691 & 0.2675 & 0.2934 & 0.7041 \\ \cmidrule(l){3-9} 
 &  & (1,3)-grams & 0.6829 & 0.2267 & 0.5643 & 0.2652 & 0.2649 & 0.6977 \\ \cmidrule(l){2-9} 
 & \multirow{3}{*}{BOW} & (1,1)-grams & 0.6465 & 0.3587 & \textbf{0.6125} & \textbf{0.3367} & \textbf{0.4942} & 0.6827 \\ \cmidrule(l){3-9} 
 &  & (1,2)-grams & 0.6682 & 0.3176 & 0.6079 & 0.3117 & 0.4504 & 0.6984 \\ \cmidrule(l){3-9} 
 &  & (1,3)-grams & 0.6740 & 0.2997 & 0.6028 & 0.3019 & 0.4303 & 0.7022 \\ \bottomrule
\end{tabular}
\end{table}

\begin{table}[]
\centering
\caption{Cross-validation NB models for Task 2}
\label{tab6:Cross-validation_NB_ models_Task_2}
\begin{tabular}{@{}ccccccccc@{}}
\toprule
NB Algorithm & Encoder & Vocabulary & Accuracy & F1-macro & F1-weighted & Recall & Precision & CEM \\ \midrule
\multirow{6}{*}{Multinomial} & \multirow{3}{*}{TF-IDF} & (1,1)-grams & 0.6743 & 0.2160 & 0.5523 & 0.2576 & 0.2393 & 0.6808 \\ \cmidrule(l){3-9} 
 &  & (1,2)-grams & \textbf{0.6769} & 0.2161 & 0.5528 & 0.2583 & 0.2417 & \textbf{0.6882} \\ \cmidrule(l){3-9} 
 &  & (1,3)-grams & 0.6766 & 0.2158 & 0.5523 & 0.2580 & 0.2686 & 0.6881 \\ \cmidrule(l){2-9} 
 & \multirow{3}{*}{BOW} & (1,1)-grams & 0.6151 & 0.2736 & 0.5806 & 0.2807 & 0.2796 & 0.6384 \\ \cmidrule(l){3-9} 
 &  & (1,2)-grams & 0.6061 & 0.2747 & 0.5798 & 0.2858 & 0.2858 & 0.6473 \\ \cmidrule(l){3-9} 
 &  & (1,3)-grams & 0.5137 & 0.2657 & 0.5250 & 0.2878 & 0.2810 & 0.6133 \\ \midrule
\multirow{6}{*}{Bernoulli} & \multirow{3}{*}{TF-IDF} & (1,1)-grams & 0.6220 & 0.2472 & 0.5491 & 0.2631 & 0.2782 & 0.6210 \\ \cmidrule(l){3-9} 
 &  & (1,2)-grams & 0.6396 & 0.2212 & 0.5451 & 0.2504 & 0.2302 & 0.6286 \\ \cmidrule(l){3-9} 
 &  & (1,3)-grams & 0.6396 & 0.2175 & 0.5426 & 0.2485 & 0.2252 & 0.6268 \\ \cmidrule(l){2-9} 
 & \multirow{3}{*}{BOW} & (1,1)-grams & 0.6220 & 0.2472 & 0.5491 & 0.2631 & 0.2782 & 0.6210 \\ \cmidrule(l){3-9} 
 &  & (1,2)-grams & 0.6396 & 0.2212 & 0.5451 & 0.2504 & 0.2302 & 0.6286 \\ \cmidrule(l){3-9} 
 &  & (1,3)-grams & 0.6396 & 0.2175 & 0.5426 & 0.2485 & 0.2252 & 0.6268 \\ \midrule
\multirow{6}{*}{Gaussian} & \multirow{3}{*}{TF-IDF} & (1,1)-grams & 0.4031 & 0.2311 & 0.4429 & 0.2345 & 0.2471 & 0.5075 \\ \cmidrule(l){3-9} 
 &  & (1,2)-grams & 0.4923 & 0.2530 & 0.5056 & 0.2540 & 0.2586 & 0.5376 \\ \cmidrule(l){3-9} 
 &  & (1,3)-grams & 0.4915 & 0.2532 & 0.5058 & 0.2541 & 0.2592 & 0.5386 \\ \cmidrule(l){2-9} 
 & \multirow{3}{*}{BOW} & (1,1)-grams & 0.4000 & 0.2333 & 0.4418 & 0.2398 & 0.2504 & 0.5081 \\ \cmidrule(l){3-9} 
 &  & (1,2)-grams & 0.4834 & 0.2534 & 0.5009 & 0.2566 & 0.2590 & 0.5361 \\ \cmidrule(l){3-9} 
 &  & (1,3)-grams & 0.4834 & 0.2538 & 0.5015 & 0.2570 & 0.2597 & 0.5370 \\ \midrule
\multirow{6}{*}{Complement} & \multirow{3}{*}{TF-IDF} & (1,1)-grams & 0.5911 & 0.2746 & 0.5648 & 0.2837 & 0.2893 & 0.5948 \\ \cmidrule(l){3-9} 
 &  & (1,2)-grams & 0.6483 & 0.2749 & 0.5811 & 0.2880 & \textbf{0.3171} & 0.6342 \\ \cmidrule(l){3-9} 
 &  & (1,3)-grams & 0.6497 & 0.2800 & \textbf{0.5845} & 0.2936 & 0.3162 & 0.6377 \\ \cmidrule(l){2-9} 
 & \multirow{3}{*}{BOW} & (1,1)-grams & 0.4932 & \textbf{0.2916} & 0.5289 & 0.3211 & 0.3002 & 0.5751 \\ \cmidrule(l){3-9} 
 &  & (1,2)-grams & 0.3647 & 0.2509 & 0.4386 & \textbf{0.3365} & 0.3035 & 0.5459 \\ \cmidrule(l){3-9} 
 &  & (1,3)-grams & 0.2010 & 0.1749 & 0.2712 & 0.3261 & 0.3086 & 0.4923 \\ \bottomrule
\end{tabular}
\end{table}

Tables \ref{tab7:top_5_mBERT_model_cross-validation_Task_1}, \ref{tab8:top_5_BETO_model_cross-validation_Task_1}, \ref{tab9:top_5_mBERT_model_cross-validation_Task_2}, and \ref{tab10:top_5_BETO_model_cross-validation_Task_2} show the top 5 results obtained in the 10-fold cross-validation process for the BERT models.  We used the DETOXIS official metrics to rank the models, which are the F1-score for Task 1 and the CEM for Task 2. Tables \ref{tab7:top_5_mBERT_model_cross-validation_Task_1} and \ref{tab8:top_5_BETO_model_cross-validation_Task_1}, in this sequence,  show the top 5 results of the mBERT model and the BETO model for Task 1. Tables \ref{tab9:top_5_mBERT_model_cross-validation_Task_2} and \ref{tab10:top_5_BETO_model_cross-validation_Task_2} respectively show the top 5 results of the mBERT model and the BETO model for Task 2.  In all four tables, the first column shows the BERT model, and the second column displays the type of Output BERT. The third column shows Learning Rate, the fourth column shows the Batch Size, and the fifth column indicates the number of Epochs. The rest of the columns have the evaluation metrics for each group of the selected parameters. For Task 1, the evaluation metrics are Accuracy, F1-score, Recall, and Precision, and for Task 2, the evaluation metrics are Accuracy, F1-macro, F1-weighted, Recall, Precision, and CEM.

\begin{table}[]
\centering
\caption{Top 5 mBERT models cross-validation for Task 1}
\label{tab7:top_5_mBERT_model_cross-validation_Task_1}
\begin{tabular}{@{}ccccccccc@{}}
\toprule
Model & \begin{tabular}[c]{@{}c@{}}Output \\ BERT\end{tabular} & \begin{tabular}[c]{@{}c@{}}Learning \\ Rate\end{tabular} & \begin{tabular}[c]{@{}c@{}}Batch \\ Size\end{tabular} & Epochs & Accuracy & F1-score & Recall & Precision \\ \midrule
\multirow{5}{*}{mBERT} & pooler & 3E-05 & 32 & 11 & 0.6972 & \textbf{0.6010} & 0.6842 & 0.5594 \\ \cmidrule(l){2-9} 
 & hidden & 5E-05 & 32 & 8 & 0.7094 & 0.5865 & 0.6167 & 0.5759 \\ \cmidrule(l){2-9} 
 & hidden & 5E-05 & 32 & 9 & 0.7102 & 0.5838 & 0.6202 & 0.5713 \\ \cmidrule(l){2-9} 
 & hidden & 3E-05 & 64 & 16 & 0.7259 & 0.5819 & 0.5778 & 0.6083 \\ \cmidrule(l){2-9} 
 & pooler & 3E-05 & 16 & 8 & 0.6838 & 0.5798 & 0.6715 & 0.5319 \\ \bottomrule
\end{tabular}
\end{table}

\begin{table}[]
\centering
\caption{Top 5 BETO models cross-validation for Task 1}
\label{tab8:top_5_BETO_model_cross-validation_Task_1}
\begin{tabular}{@{}ccccccccc@{}}
\toprule
Model & \begin{tabular}[c]{@{}c@{}}Output \\ BERT\end{tabular} & \begin{tabular}[c]{@{}c@{}}Learning \\ Rate\end{tabular} & \begin{tabular}[c]{@{}c@{}}Batch \\ Size\end{tabular} & Epochs & Accuracy & F1-score & Recall & Precision \\ \midrule
\multirow{5}{*}{BETO} & pooler & 1E-05 & 32 & 4 & 0.7446 & \textbf{0.6314} & 0.6514 & 0.6338 \\ \cmidrule(l){2-9} 
 & pooler & 1E-05 & 64 & 7 & 0.7415 & 0.6276 & 0.6578 & 0.6184 \\ \cmidrule(l){2-9} 
 & pooler & 1E-05 & 16 & 7 & 0.7267 & 0.6265 & 0.6829 & 0.6016 \\ \cmidrule(l){2-9} 
 & pooler & 5E-05 & 64 & 14 & 0.7554 & 0.6245 & 0.6203 & 0.6485 \\ \cmidrule(l){2-9} 
 & pooler & 3E-05 & 64 & 16 & 0.7565 & 0.6237 & 0.6117 & 0.6568 \\ \bottomrule
\end{tabular}
\end{table}

\begin{table}[]
\centering
\caption{Top 5 mBERT models cross-validation for Task 2}
\label{tab9:top_5_mBERT_model_cross-validation_Task_2}
\begin{tabular}{@{}ccccccccccc@{}}
\toprule
Model & \begin{tabular}[c]{@{}c@{}}Output \\ BERT\end{tabular} & \begin{tabular}[c]{@{}c@{}}Learning \\ Rate\end{tabular} & \begin{tabular}[c]{@{}c@{}}Batch \\ Size\end{tabular} & Epochs & Accuracy & \begin{tabular}[c]{@{}c@{}}F1\\ macro\end{tabular} & \begin{tabular}[c]{@{}c@{}}F1\\ weighted\end{tabular} & Recall & Precision & CEM \\ \midrule
\multirow{5}{*}{mBERT} & pooler & 1E-05 & 16 & 12 & 0.7031 & 0.4165 & 0.7477 & 0.4206 & 0.4483 & \textbf{0.7599} \\ \cmidrule(l){2-11} 
 & hidden & 1E-05 & 16 & 14 & 0.6955 & 0.4158 & 0.7486 & 0.4252 & 0.4344 & 0.7588 \\ \cmidrule(l){2-11} 
 & hidden & 3E-05 & 16 & 4 & 0.7006 & 0.3839 & 0.7475 & 0.3970 & 0.4054 & 0.7581 \\ \cmidrule(l){2-11} 
 & hidden & 1E-05 & 16 & 10 & 0.7011 & 0.3832 & 0.7515 & 0.3917 & 0.4210 & 0.7580 \\ \cmidrule(l){2-11} 
 & pooler & 1E-05 & 16 & 4 & 0.6974 & 0.3984 & 0.7496 & 0.4067 & 0.4182 & 0.7580 \\ \bottomrule
\end{tabular}
\end{table}

\begin{table}[]
\centering
\caption{Top 5 BETO models cross-validation for Task 2}
\label{tab10:top_5_BETO_model_cross-validation_Task_2}
\begin{tabular}{@{}ccccccccccc@{}}
\toprule
Model & \begin{tabular}[c]{@{}c@{}}Output \\ BERT\end{tabular} & \begin{tabular}[c]{@{}c@{}}Learning \\ Rate\end{tabular} & \begin{tabular}[c]{@{}c@{}}Batch \\ Size\end{tabular} & Epochs & Accuracy & \begin{tabular}[c]{@{}c@{}}F1\\ macro\end{tabular} & \begin{tabular}[c]{@{}c@{}}F1\\ weighted\end{tabular} & Recall & Precision & CEM \\ \midrule
\multirow{5}{*}{BETO} & hidden & 1E-05 & 16 & 4 & 0.7170 & 0.4035 & 0.7678 & 0.4091 & 0.4469 & \textbf{0.7769} \\ \cmidrule(l){2-11} 
 & hidden & 1E-05 & 8 & 3 & 0.7165 & 0.4138 & 0.7696 & 0.4151 & 0.4611 & 0.7747 \\ \cmidrule(l){2-11} 
 & hidden & 3E-05 & 32 & 6 & 0.7188 & 0.4096 & 0.7483 & 0.4173 & 0.4355 & 0.7746 \\ \cmidrule(l){2-11} 
 & hidden & 3E-05 & 64 & 5 & 0.7148 & 0.4178 & 0.7632 & 0.4235 & 0.4461 & 0.7746 \\ \cmidrule(l){2-11} 
 & hidden & 1E-05 & 8 & 5 & 0.7153 & 0.4168 & 0.7649 & 0.4219 & 0.4592 & 0.7739 \\ \bottomrule
\end{tabular}
\end{table}

\subsection{Best model}

\begin{figure}
     \centering
    \includegraphics[scale=0.08]{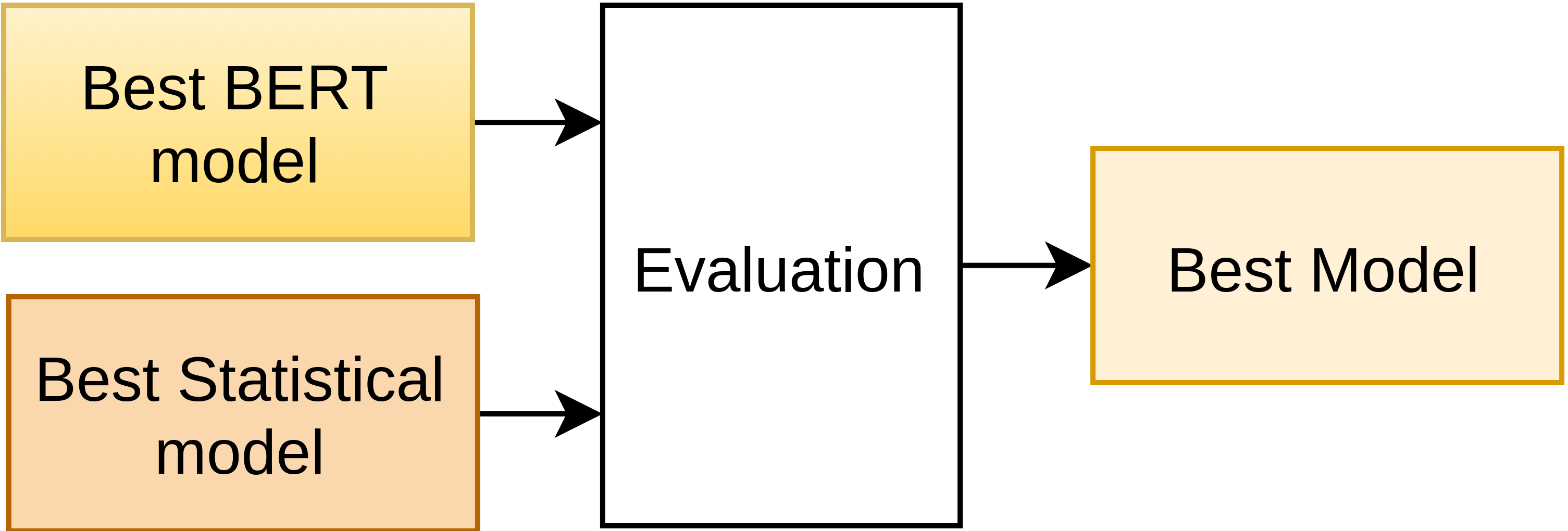}
    \caption{Selection of the best ML model.} 
    \label{fig5:Selection_of_the_best_ML_model}
\end{figure}

At the end of the cross-validation, we selected the best model for each task accordingly with the DETOXIS official metric for the specified task, as shown in Figure \ref{fig5:Selection_of_the_best_ML_model}.

Table \ref{tab11:The_best_result_of_each_model_in_the_cross-validation_Task_1} shows the best results for each ML model tried on the cross-validation process for Task 1, which are mBERT, BETO, NB, and ME. Table \ref{tab12:top_5_model_cross-validation_Task_1} displays the top 5 best model for the Task 1 during the whole cross-validation process.

\begin{table}[]
\centering
\caption{The best result of each model in the cross-validation for Task 1}
\label{tab11:The_best_result_of_each_model_in_the_cross-validation_Task_1}
\begin{tabular}{@{}ccccc@{}}
\toprule
Model & Accuracy & F1-score & Recall & Precision \\ \midrule
BETO & 0.7446 & \textbf{0.6314} & 0.6514 & 0.6338 \\
mBERT & 0.6972 & 0.6010 & 0.6842 & 0.5594 \\
NB & 0.5795 & 0.5355 & 0.7289 & 0.4238 \\
ME & 0.7002 & 0.4679 & 0.4019 & 0.5670 \\ \bottomrule
\end{tabular}
\end{table}

\begin{table}[]
\centering
\caption{Top 5 models cross-validation for Task 1}
\label{tab12:top_5_model_cross-validation_Task_1}
\begin{tabular}{@{}ccccccccc@{}}
\toprule
Model & \begin{tabular}[c]{@{}c@{}}Output \\ BERT\end{tabular} & \begin{tabular}[c]{@{}c@{}}Learning \\ Rate\end{tabular} & \begin{tabular}[c]{@{}c@{}}Batch \\ Size\end{tabular} & Epochs & Accuracy & F1-score & Recall & Precision \\ \midrule
BETO & pooler & 1E-05 & 32 & 4 & 0.7446 & \textbf{0.6314} & 0.6514 & 0.6338 \\
BETO & pooler & 1E-05 & 64 & 7 & 0.7415 & 0.6276 & 0.6578 & 0.6184 \\
BETO & pooler & 1E-05 & 16 & 7 & 0.7267 & 0.6265 & 0.6829 & 0.6016 \\
BETO & pooler & 5E-05 & 64 & 14 & 0.7554 & 0.6245 & 0.6203 & 0.6485 \\
BETO & pooler & 3E-05 & 64 & 16 & 0.7565 & 0.6237 & 0.6117 & 0.6568 \\ \bottomrule
\end{tabular}
\end{table}

Table \ref{tab13:The_best_result_of_each_model_in_the_cross-validation_Task_2} shows the best models performace for each ML model on the cross-validation process for Task 2, which are mBERT, BETO, ME, and NB. Table \ref{tab14:top_5_model_cross-validation_Task_2} displays the top 5 best model for the Task 2 whole cross-validation process.

\begin{table}[]
\centering
\caption{The best result of each model in the cross-validation for Task 2}
\label{tab13:The_best_result_of_each_model_in_the_cross-validation_Task_2}
\begin{tabular}{@{}ccccccc@{}}
\toprule
Model & Accuracy & F1-macro & F1-weighted & Recall & Precision & CEM \\ \midrule
BETO & 0.7170 & 0.4035 & 0.7678 & 0.4091 & 0.4469 & \textbf{0.7769} \\
mBERT & 0.7031 & 0.4165 & 0.7477 & 0.4206 & 0.4483 & 0.7599 \\
ME & 0.6780 & 0.2944 & 0.5998 & 0.2995 & 0.4418 & 0.7080 \\
NB & 0.6769 & 0.2161 & 0.5528 & 0.2583 & 0.2417 & 0.6882 \\ \bottomrule
\end{tabular}
\end{table}

\begin{table}[H]
\centering
\caption{Top 5 models cross-validation for Task 2}
\label{tab14:top_5_model_cross-validation_Task_2}
\begin{tabular}{@{}ccccccccccc@{}}
\toprule
Model & \begin{tabular}[c]{@{}c@{}}Output \\ BERT\end{tabular} & \begin{tabular}[c]{@{}c@{}}Learning \\ Rate\end{tabular} & \begin{tabular}[c]{@{}c@{}}Batch \\ Size\end{tabular} & Epochs & Accuracy & \begin{tabular}[c]{@{}c@{}}F1\\ macro\end{tabular} & \begin{tabular}[c]{@{}c@{}}F1\\ weighted\end{tabular} & Recall & Precision & CEM \\ \midrule
BETO & hidden & 1E-05 & 16 & 4 & 0.7170 & 0.4035 & 0.7678 & 0.4091 & 0.4469 & \textbf{0.7769} \\
BETO & hidden & 1E-05 & 8 & 3 & 0.7165 & 0.4138 & 0.7696 & 0.4151 & 0.4611 & 0.7747 \\
BETO & hidden & 3E-05 & 32 & 6 & 0.7188 & 0.4096 & 0.7483 & 0.4173 & 0.4355 & 0.7746 \\
BETO & hidden & 3E-05 & 64 & 5 & 0.7148 & 0.4178 & 0.7632 & 0.4235 & 0.4461 & 0.7746 \\
BETO & hidden & 1E-05 & 8 & 5 & 0.7153 & 0.4168 & 0.7649 & 0.4219 & 0.4592 & 0.7739 \\ \bottomrule
\end{tabular}
\end{table}

After the cross-validation, we chose the best model for Task 1, which following Table \ref{tab12:top_5_model_cross-validation_Task_1} is BETO with the respective parameters: (i) pooler as Output BERT; (ii) 1E-05 Learning Rate; (iii) Batch Size equal 32; and (iv) 4 training Epochs. We also selected the best model for Task 2 that following Table \ref{tab14:top_5_model_cross-validation_Task_2} is BETO with the respective parameters: (i) hidden Output BERT; (ii) 1E-05 Learning Rate; (iii) Batch Size equal 16; and (iv) 4 training Epochs. Having the best models and their parameters, we trained the models on the train set.


Once the best models are trained, we use those models to make the predictions on the DETOXIS test set. These predictions afterward were submitted to the DETOXIS shared task organization as our final results.


\section{Results and Discussion}

We discovered important information on the cross-validation results. Looking at Table \ref{tab3:Cross-validation_ME_models_Task_1}, we can see that the ME model achieves its best results on Task 1 with the BOW encode based on the F1-score evaluation metric, which is 0.4679. The highest Accuracy 0.7126 and Recall 0.4019 are also performed with the BOW encode. The only performance metric in which the TF-IDF encode obtains a higher score is Precision that is 0.8928. Thus, we can conclude that BOW is the best encoding for the ME model on Task 1 in the DETOXIS training set. Moreover, employing the Sag solver, the ME model achieved a higher F1-score, Recall, and Precision.  Hence, it seems to us that Sag was the best solver for the ME model on Task 1 in the DETOXIS training set. We do not have a definitive conclusion about the vocabulary size because the ME model achieved its highest results with different numbers of n-grams for each metric.

Observing Table \ref{tab4:Cross-validation_NB_models_Task_1}, we see that the NB model achieves its best results on Task 1 with the BOW encode based on the F1-score evaluation metric, which is 0.5355. The NB model also obtained the higest Recall 0.8004 with the BOW encode, but its highest results for Accuracy 0.6933 and Precision 0.7282 were with the TF-IDF encode. Therefore, we can not conclude which encode method is the best for the NB model on Task 1 in the DETOXIS training set. A similar case occurs with the vocabulary size where the NB model that employed 1-grams, 2-grams, and 3-grams achieved the highest  F1-score and Recall. However, the NB model with a 1-grams vocabulary size obtained the highest Accuracy and Precision. The different NB algorithms obtained a similar performance based on the F1-score. In most cases, they achieved their best results with the BOW encode.

We can see in Table \ref{tab5:Cross-validation_ME_models_Task_2} that the ME model achieved its best results on Task 2 with the BOW encode based on the CEM evaluation metric, which is 0.7080. The highest F1-macro 0.3587, F1-weighted 0.6125, Recall 0.3367, and Precision 0.4942 are also obtained with the BOW encode. The only performance metric in which the TF-IDF encode obtains a higher score is the Accuracy, which is 0.6846.  Thus, we can conclude that BOW is the best encoding for the ME model on Task 2 in the DETOXIS training set. Moreover, employing the Newton solver, the ME model achieved a higher Accuracy, F1-macro, F1-weighted, Recall, and Precision.  Hence, it seems to us that Newton was the best solver for the ME model on Task 2 in the DETOXIS training set. We concluded that the vocabulary size of 1-grams is the best for the ME model on Task 2 in the DETOXIS training set because the ME model achieved its highest Accuracy, F1-macro, F1-weighted, Recall, and Precision.

Table \ref{tab6:Cross-validation_NB_ models_Task_2} shows that the NB model achieved its best results on Task 2 with the TF-IDF encode based on the CEM evaluation metric, which is 0.6882. The NB model also obtained its highest Accuracy 0.6769, F1-weighted 0.5845, and Precision 0.3171, with the TF-IDF encode, but its highest results for F1-macro 0.2916 and Recall 0.3365 were obtained with the BOW encode. Therefore, we can conclude that the TF-IDF encode best suits the NB model on Task 2 in the DETOXIS training set. We see indications in Table \ref{tab6:Cross-validation_NB_ models_Task_2} that the ideal vocabulary for the NB model on Task 2 in the DETOXIS training set is composed of 1-grams and 2 grams. Once with this vocabulary, the model obtained its highest Accuracy, Recall, Precision, and CEM results. The different NB algorithms obtained similar performance based on the CEM ranged from 0.49 to 0.68.  

Based on the F1-score, the mBERT model achieved its best performance on Task 1 with a value of 0.6010, as we can see in Table \ref{tab7:top_5_mBERT_model_cross-validation_Task_1}. The model parameters are the following: (i) pooler as Output BERT; (ii) 3E-05 Learning Rate; (iii) Batch Size equal 32; and (iv) 11 training Epochs. Table \ref{tab8:top_5_BETO_model_cross-validation_Task_1} shows that the BETO model obtained its best performance on Task 1
also based on the F1-score with the following parameters: (i) pooler as Output BERT; (ii) 1E-05 Learning Rate; (iii) Batch Size equal 32; and (iv) 4 training Epochs. The BETO model obtained a F1-score value of 0.6314, which was also the highest among all the ML models in the cross-validation process. For this reason, the BETO model with the mentioned parameters was used for our Task 1 official prediction on the DETOXIS test set. These predictions afterward were submitted as our official Task 1 results.

Observing Table \ref{tab9:top_5_mBERT_model_cross-validation_Task_2}, we can conclude that based on the CEM, the mBERT model achieved its best performance on Task 1 with the following parameters: (i) pooler as Output BERT; (ii) 1E-05 Learning Rate; (iii) Batch Size equal to 16; and (iv) 12 training Epochs. This model achieved the CEM of 0.7599. Table \ref{tab10:top_5_BETO_model_cross-validation_Task_2} shows that the BETO model obtained its best performance on Task 2 also based on the CEM with the following parameters: (i) hidden as Output BERT; (ii) 1E-05 Learning Rate; (iii) Batch Size equal to 16; and (iv) 4 training Epochs. The BETO model obtained CEM value of 0.7769, which was also the highest among all the ML models in the cross-validation process. For this reason, the BETO model with the mentioned parameters was used for our Task 2 official prediction on the DETOXIS test set. These predictions afterward were submitted as our official Task 2 results.

To sum up the comments about the cross-validation results, looking at Tables \ref{tab12:top_5_model_cross-validation_Task_1} and \ref{tab14:top_5_model_cross-validation_Task_2}, we can see that the BETO model with different combinations of parameters obtained the five first positions on the ranking for the best ML model for Task 1 and Task 2.

The DETOXIS organization provided us with the results of the test set. Table \ref{tab15:Test_set_results_for_Task_1} shows our result on Task 1 plus the three official DETOXIS baselines: Random Classifier, Chain BOW, and BOW Classifier.  Our model obtained an F1-score around 59\% greater than the results obtained by the best baseline on Task 1.

\begin{table}[]
\centering
\caption{Test set results for Task 1}
\label{tab15:Test_set_results_for_Task_1}
\begin{tabular}{@{}cc@{}}
\toprule
Model & F1-score \\ \midrule
\textbf{BETO} & \textbf{0.5996} \\
Random Classifier & 0.3761 \\
Chain BOW & 0.3747 \\
BOW Classifier & 0.1837 \\ \bottomrule
\end{tabular}
\end{table}

Table \ref{tab16:Test_set_results_for_Task_2} shows the results of our model and the three DETOXIS baselines on Task 2. Our BETO model was able to achieve a CEM of 9\% higher than the best DETOXIS baseline result obtained by the Random Classifier.

\begin{table}[]
\centering
\caption{Test set results for Task 2}
\label{tab16:Test_set_results_for_Task_2}
\begin{tabular}{@{}cc@{}}
\toprule
Model & CEM \\ \midrule
\textbf{BETO} & \textbf{0.7142} \\
Chain BOW & 0.6535 \\
BOW Classifier & 0.6318 \\
Random Classifier & 0.4382 \\ \bottomrule
\end{tabular}
\end{table}

On the DETOXIS official ranking, we obtained 3rd place on Task 1 with F1-score 0.5996, and we achieved 6th place on Task 2 with CEM 0.7142.

\section{Conclusion and Future Work}

Xenophobia is a problem which is aggravated by the increase in the spread of toxic comments posted in different online news articles related to immigration. In this paper, to address this problem within the DETOXIS 2021 shared task, we tried two types of ML models: (i) statistical models and (ii) BERT models. We obtained the best results in both tasks using BETO, a BERT model pre-trained with a big Spanish corpus. Our contributions are as follows: (i) help in the effort to improve the results in the identification of toxic comments in news articles related to immigration. Unlike the vast majority of works, we use ML models that can tackle the xenophobia detection problem having only little data available;  (ii)  We build an ML model and find its best configuration to deal not only with the classification of news articles as `toxic' and `not toxic', but also to infer the toxicity level of the comments into `not toxic', `mildly toxic', `toxic', or `very toxic'.  

Based on the DETOXIS official metrics, we concluded that our results indicate that: (i) BERT models obtain better results than statistical models for toxicity and toxicity level detection in text comments; and (ii) Monolingual BERT models achieve higher results in comparison with the multilingual BERT models in toxicity detection and toxicity level detection in their pre-trained language.

After all, our BETO model obtained the 3rd position on Task 1 official ranking with the F1-score of 0.5996, and it achieved the 6th position on Task 2 official ranking with the CEM of 0.7142. As future work, we aim to include sentiment lexicons on the model's input to boost its performance.

%
%
%
\bibliographystyle{splncs04}
\bibliography{bibliography}
%




\end{document}